\documentclass[journal]{IEEEtran}
\usepackage{amsmath,amsfonts}
\usepackage{algorithmic}
\usepackage{algorithm}
\usepackage{array}
\usepackage[caption=false,font=normalsize,labelfont=sf,textfont=sf]{subfig}
\usepackage{textcomp}
\usepackage{stfloats}
\usepackage{url}
\usepackage{verbatim}
\usepackage{graphicx}
\usepackage[numbers]{natbib}
\usepackage{authblk}
\usepackage{hyperref}
\usepackage{booktabs}
\usepackage{tabularx}
\usepackage{colortbl}
\usepackage{multirow}
\usepackage{adjustbox}
\usepackage[table,xcdraw]{xcolor}
\usepackage[capitalise,noabbrev]{cleveref}
\usepackage[utf8]{inputenc}
\usepackage{forest}
\usepackage{booktabs}
\usepackage{enumitem}
\usepackage{float} 
\usepackage{subfig}
\def\BibTeX{{\rm B\kern-.05em{\sc i\kern-.025em b}\kern-.08em
    T\kern-.1667em\lower.7ex\hbox{E}\kern-.125emX}}
\usepackage{balance}
\renewcommand{\thetable}{\Roman{table}}

\begin{document}

\title{PPTNet: A Hybrid Periodic Pattern-Transformer Architecture for Traffic Flow Prediction and Congestion Identification}

\markboth{SUBMISSION TO IEEE TRANSACTIONS ON INTELLIGENT TRANSPORTATION SYSTEMS}%
         {}

\author{Hongrui Kou\thanks{This work was supported in part by the Science and Technology Development Program of Jilin Province under Grant 20240302052GX and in part by the National Natural Science Foundation of China under Grant 52075213.}\thanks{Hongrui Kou, Ziyu Wang, and Zhouhang Lv are with National Key Laboratory of Automotive Chassis Integration and Bionics, Jilin University, Changchun 130000, China (email: kouhr23@mails.jlu.edu.cn; zyw22@mails.jlu.edu.cn;  lvzh22@mails.jlu.edu.cn).}, Jingkai Li\thanks{Jingkai Li is with School of Mechanical Engineering, Shandong University, Jinan 250061, China (email: 202334397@mail.sdu.edu.cn).}, Ziyu Wang, Zhouhang Lv, Yuxin Zhang* \thanks{* Yuxin Zhang is the corresponding author and also with National Key Laboratory of Automotive Chassis Integration and Bionics, Jilin University, Changchun 130000, China (email: yuxinzhang@jlu.edu.cn).}, Cheng Wang\thanks{Cheng Wang is with School of Engineering and Physical Sciences, Heriot-Watt University, Edinburgh EH1, United Kingdom (email: Cheng.Wang@hw.ac.uk).}}


\maketitle
\begin{abstract}
Accurate prediction of traffic flow parameters and real-time identification of congestion states are essential for the efficient operation of intelligent transportation systems. This paper proposes a Periodic Pattern-Transformer Network (PPTNet) for traffic flow prediction, integrating periodic pattern extraction with the Transformer architecture, coupled with a fuzzy inference method for real-time congestion identification. Firstly, a high-precision traffic flow dataset (Traffic Flow Dataset for China’s Congested Highways \& Expressways, TF4CHE) suitable for congested highway scenarios in China is constructed based on drone aerial imagery data. Subsequently, the proposed PPTNet employs Fast Fourier Transform to capture multi-scale periodic patterns and utilizes two-dimensional Inception convolutions to efficiently extract intra and inter periodic features. A Transformer decoder dynamically models temporal dependencies, enabling accurate predictions of traffic density and speed. Finally, congestion probabilities are calculated in real-time using the predicted outcomes via a Mamdani fuzzy inference-based congestion identification module. Experimental results demonstrate that the proposed PPTNet significantly outperforms mainstream traffic prediction methods in prediction accuracy, and the congestion identification module effectively identifies real-time road congestion states, verifying the superiority and practicality of the proposed method in real-world traffic scenarios. Project page: \href{https://github.com/ADSafetyJointLab/PPTNet}{https://github.com/ADSafetyJointLab/PPTNet}.

\end{abstract}


\section{Introduction}
\begin{figure*}[htbp]
    \centering
    \includegraphics[width=1\linewidth]{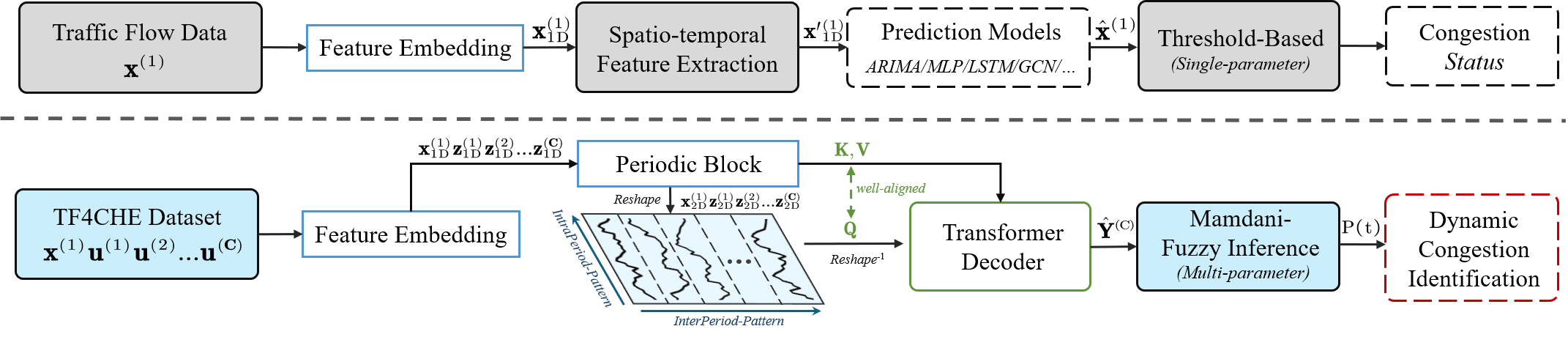}
    \caption{(a) The traditional method uses a single data source for predicting traffic flow data and identifies congestion through a threshold-based single- parameter method; (b) We propose a novel framework for traffic prediction and congestion dentification, which takes multidimensional traffic flow data $\left\{\mathbf{u}^{(1)},\;\mathbf{u}^{(2)},\;\dots,\;\mathbf{u}^\mathbf{(C)}\right\}$ as auxiliary input, uses periodic blocks to achieve adaptive periodic feature mining and uses transformer decoder to achieve prediction output. Combined with a multi-parameter Mamdani fuzzy inference mechanism for real-time and accurate congestion identification.}
    \label{fig: pipline}
\end{figure*}

\IEEEPARstart{W}{ith} the rapid advancement of Intelligent Transportation Systems (ITS), traffic flow prediction has become a core technology to optimize traffic management and improve operational efficiency \cite{shao2024adaptive}. As a critical component of national transportation infrastructure, expressways are particularly susceptible to traffic congestion, which not only directly reduces throughput but also indirectly contributes to a higher incidence of traffic accidents. Therefore, enabling congestion identification based on traffic flow prediction results is of great importance for alleviating congestion and reducing accident risks. However, urgent challenges remain in expressway traffic data acquisition, accurate traffic flow prediction, and effective congestion identification. \hyperref[fig: pipline]{Fig.1} illustrates the limitations of current traffic congestion prediction. Therefore, how to effectively integrate traffic flow prediction with congestion identification techniques to achieve real-time congestion warning and response has emerged as a key research focus in the field of intelligent transportation.  

Traffic flow data perception and dataset acquisition must meet the demands of high accuracy and comprehensive temporal coverage in complex road network conditions. Traditional traffic flow data are primarily collected using fixed-location methods, including stationary sensors \cite{chen2001freeway}, fixed surveillance cameras \cite{yu2019citywide}, and ultrasonic detectors \cite{li2019detection}. These methods exhibit several limitations, including high measurement errors, limited spatial coverage, and substantial deployment costs. In recent years, mobile data collection techniques have undergone significant advancements. In particular, unmanned aerial vehicle (UAV) remote sensing—owing to its advantages in high resolution, multi-dimensionality, and dynamic observation—has emerged as a novel solution for real-time and efficient traffic data acquisition \cite{krajewski2018highd}. Moreover, The high-resolution, multi-dimensional, and dynamically adaptive sensing capabilities of UAV-based remote sensing technologies have further opened new avenues for traffic flow modeling.

Traffic flow prediction is inherently a nonlinear time-series forecasting problem \cite{medina2022urban}, with core challenges centered on modeling its highly dynamic and periodic characteristics accurately. The Greenshields model, proposed in 1935, revealed complex phase transition relationships among key traffic parameters such as flow, density, and speed \cite{greenshields1935study}. Early studies based on the Seasonal Autoregressive Integrated Moving Average (SARIMA) model \cite{kumar2015short} achieved moderate success in capturing periodic patterns, but exhibited limited capability in modeling nonlinear dynamics. With the emergence of machine learning techniques, algorithms such as Support Vector Regression (SVR) \cite{hu2014traffic} and Gradient Boosting Decision Trees (GBDT) \cite{cheng2018research} have demonstrated promising nonlinear fitting capabilities. However, they still fall short of capturing deep temporal dependencies. In recent years, deep learning-based approaches have become dominant. Methods such as Convolutional Neural Networks (CNNs) \cite{zeng2021modeling}, Recurrent Neural Networks (RNNs) \cite{belhadi2020recurrent}, Long Short-Term Memory (LSTM) networks \cite{yang2019traffic}, Graph Convolutional Networks (GCNs) \cite{yu2017spatio}, and Transformer models \cite{vaswani2017attention} have been widely applied in traffic flow prediction. Although these methods have improved prediction performance to some extent, they still suffer from issues such as gradient vanishing and difficulty in modeling long-term dependencies—especially when dealing with highly nonlinear and strongly periodic traffic flow data, resulting in prediction outcomes that often remain suboptimal.

Effective congestion identification requires accounting for the dynamic nature of traffic flow. A critical challenge lies in quantitatively characterizing the evolving dynamics of traffic to provide a reliable foundation for congestion detection \cite{liu2013traffic}. Most existing congestion identification methods rely on fixed thresholds or single-parameter criteria, which exhibit poor adaptability and high subjectivity \cite{drliciak2024research}. Although some studies have introduced fuzzy logic systems to reduce subjectivity in the identification process, these methods typically consider only a single variable, such as speed or density, as the decision basis \cite{alkaissi2024traffic}. Moreover, current congestion identification approaches are predominantly retrospective in nature, lacking seamless integration with traffic flow prediction technologies. As a result, they fail to meet the practical requirements of real-time congestion warning and early response.

To address the appeal challenges, this paper proposes a highway traffic flow prediction and congestion identification framework based on a Periodic Pattern-Transformer Network (PPTNet). The contribution of this paper is summarized as follows:

\begin{enumerate}[label=\arabic*)]
    \item Leveraging the drone aerial driving dataset, high-precision multidimensional traffic flow features are extracted to develop TF4CHE, a specialized traffic flow dataset suitable for China’s Congested Highways \& Expressways.
    \item Propose a PPTNet short-term traffic flow prediction model, which adaptively captures multi period features in multidimensional traffic data through fast fourier transform and effectively extracts inter period and intra period feature pattern  through two-dimensional Inception convolution; Subsequently, the Transformer decoder is used to dynamically model short-term nonlinear changes, achieving accurate prediction of future traffic conditions.
    \item Build a  Mamdani fuzzy inference module that maps traffic density and average speed to continuous congestion probability. This module uses a Gaussian membership function to fuzzy the input into three levels, generates four levels of congestion intensity using nine double-input single-output rules, and defuzzifies the results using the center of gravity method to obtain 0–1 probability values.
\end{enumerate}

The structure of this article is as follows: Section \ref{rel} reviews related works on time-series traffic flow prediction, congestion identification, and existing traffic flow datasets. Section \ref{data} presents the preprocessing and feature extraction procedures applied to the AD4CHE dataset, through which the TF4CHE dataset—representing congested highway traffic flow in China—is constructed, along with a descriptive analysis of the extracted data. Sections \ref{PPT} and \ref{FI} describe the architecture of the proposed PPTNet model and the Mamdani fuzzy inference-based congestion identification mechanism, respectively. Section \ref{ex} validates the effectiveness of the proposed prediction model and congestion identification method using experiments conducted on the TF4CHE dataset. Finally, Section \ref{CL} concludes the paper and outlines potential directions for future research.

\section{Relate Work\label{rel}}
\subsection{Traffic Flow Prediction}
Traffic flow prediction methodologies have evolved significantly over time, progressing from traditional statistical models to advanced deep learning approaches. Early statistical methods such as ARIMA and its variants \cite{kumar2015short}, Vector Autoregression \cite{chandra2009predictions}, and Kalman filtering \cite{okutani1984dynamic} provided foundational frameworks but struggled with capturing non-linear relationships in traffic data. Machine learning approaches subsequently emerged, with SVR \cite{hu2014traffic}, Random Forests \cite{cheng2018research}, and \textit{k-NN} \cite{cai2016spatiotemporal} demonstrating improved capabilities in modeling non-linear patterns. However, these methods often failed to adequately capture spatial correlations between road segments. Deep learning models have revolutionized traffic prediction through their ability to model complex spatiotemporal dependencies. Recurrent architectures including LSTM \cite{ma2015long}, GRU \cite{fu2016using}, and Bi-LSTM \cite{yang2019traffic} excel at capturing temporal dependencies but face challenges with long-range temporal modeling. Convolution-based methods like TCN \cite{bai2018empirical} and attention-based models like Informer \cite{zhou2021informer} have addressed some limitations by enabling more efficient parallel processing and selective focus on relevant information; however, they remain insufficient for modeling highly nonlinear and strongly periodic traffic flow data. To better capture spatial relationships in traffic networks, ST-GCN \cite{yu2017spatio} and Graph WaveNet \cite{wu2019graph} integrate spatial and temporal dependencies through graph convolutions and adaptive adjacency matrices, while TimesNet \cite{wu2022timesnet} transforms one-dimensional time series into multiple two-dimensional tensors to capture both intra-period and inter-period variations. Despite these advances, accurately modeling complex temporal patterns and long-period features in congested traffic scenarios remains challenging.

\subsection{Traffic Congestion Identification}
Traffic congestion analysis has progressed from simplistic threshold-based methods to sophisticated multi-faceted approaches. Traditional threshold-based techniques \cite{liu2013traffic} applied predefined values to traffic parameters but lacked adaptability to varying conditions. Index-based methods including Traffic Congestion Index \cite{chen2004freeway} and Level of Service classifications \cite{elefteriadou2016highway} integrated multiple parameters but required careful calibration for specific environments. Pattern recognition techniques including various clustering algorithms have enabled identification of recurrent congestion patterns and hotspots \cite{tang2018lane}. Fuzzy logic systems have emerged as particularly valuable tools for congestion identification due to their ability to handle uncertainty in traffic parameters. Machine learning approaches including ensemble methods \cite{cheng2018research} enhance congestion prediction through feature fusion and model combination strategies, effectively integrating diverse traffic indicators from heterogeneous data sources \cite{jiang2004wavelet}. Deep learning architectures \cite{ma2015long} use specialized neural network structures to model the underlying spatiotemporal dependencies in traffic data, capturing the complex evolution patterns of congestion formation across road networks and enabling more accurate forecasting of congestion transitions. Recent innovations in hybrid approaches \cite{zhang2011data} that integrate fuzzy logic with deep learning offer promising directions by combining the interpretability of fuzzy systems with the predictive power of neural networks. The aforementioned methods have progressed from single-parameter to multi-parameter collaborative approaches in congestion detection, yet they still fail to integrate traffic flow forecasting techniques, making it difficult to satisfy the requirements for real-time congestion identification.

\subsection{Traffic Flow Datasets}
Traffic flow datasets serve as critical resources for developing and evaluating prediction methodologies, with variations in collection methods and coverage providing different perspectives on traffic dynamics. Traditional stationary sensor datasets like PeMS \cite{chen2001freeway} provide continuous measurements from fixed locations, while mobile sensor-based collections such as the T-Drive Dataset \cite{yuan2011driving} offer trajectory data from moving vehicles. Simulation-generated datasets from platforms like SUMO \cite{lopez2018microscopic} provide controlled environments for algorithm testing. Aerial and satellite imagery datasets have emerged as particularly valuable resources, offering comprehensive bird's-eye views of traffic conditions. The NGSIM Dataset \cite{punzo2011assessment} and drone-based collections including the HighD Dataset \cite{krajewski2018highd} and its variants \cite{bock2020ind} provide high-precision vehicle trajectories enabling fine-grained analysis of traffic dynamics. Chinese traffic datasets have gained increasing attention due to the unique characteristics of Chinese traffic patterns. The China-FOT Dataset \cite{liu2017driving} and SIND Dataset \cite{xu2022drone} provide valuable insights but typically focus on specific scenarios rather than congested highway conditions. The AD4CHE Dataset \cite{zhang2023ad4che} addresses this gap by providing comprehensive aerial survey data from four Chinese cities specifically focused on congestion scenarios. Building upon AD4CHE, our proposed TF4CHE (Traffic flow for China’s Congested Highways \& Expressways) dataset enhances analysis capabilities by extracting time-series data of traffic density and average speed, facilitating development of prediction methods tailored for Chinese traffic conditions.

\section{TF4CHE DATASET\label{data}}
\begin{figure}[ht]
    \centering
    \includegraphics[width=\linewidth]{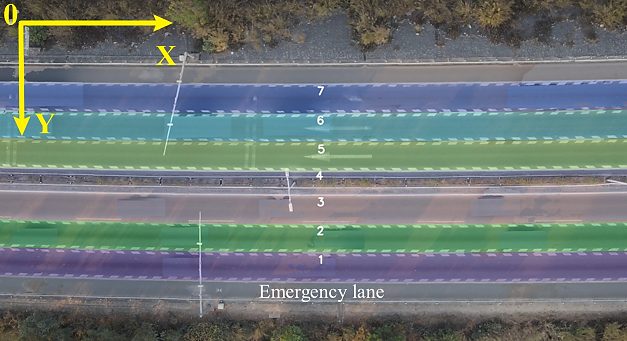}
    \caption{Lane layout of the AD4CHE dataset.}
    \label{fig: lane layout}
\end{figure}

The dataset used in this study is sourced from the Aerial Dataset for China's Congested Highways \& Expressways (AD4CHE), a specialized natural driving dataset developed by Zhang et al \cite{zhang2023ad4che}. AD4CHE focuses on congested scenarios in four cities in China and contains 5.12 hours of high-resolution aerial imagery, with an average vehicle speed of 29.25 km/h, which is significantly lower than similar datasets such as HighD, which has an average speed of 100.67 km/h. Data in AD4CHE are collected using drones at an altitude of 100 meters, employing advanced calibration techniques to ensure positioning accuracy of approximately 5 cm. The dataset provides comprehensive vehicle metrics, including position, speed, classification, and unique parameters such as self-offset and yaw rate. It includes data from expressways and arterial roads across 11 distinct road segments in five cities in China, comprising a total of 68 data segments. Each data segment consists of three files: \texttt{xx\_recordingMeta.csv} , \texttt{xx\_tracks.csv}, and \texttt{xx\_tracksMeta.csv}, which contain video metadata, vehicle trajectories, and trajectory metadata information, respectively. The lane layout for the recorded segments is shown in \hyperref[fig: lane layout]{Fig.2}. The world coordinate system has its origin at the top-left corner of the image, with the X-axis corresponding to the direction of vehicle movement and increasing to the right, while the Y-axis increases downward. There are four lanes traveling in the positive direction along the X-axis, numbered "1-4", and one emergency lane, while the lanes in the negative direction along the X-axis include "5-7" and one emergency lane.

\subsection{Data Extraction}
Based on the AD4CHE dataset, this study has constructed the Traffic Flow Dataset for China’s Congested Highways and Expressways (TF4CHE). The dataset is divided into 11 groups according to different recording road segments. It primarily extracts key time-series traffic data from the AD4CHE dataset and calculates typical traffic flow parameters using standard formulas. This dataset can be applied in various intelligent transportation applications, including traffic flow prediction, congestion identification, and other related areas.

The TF4CHE dataset includes various typical time-series traffic flow parameters, such as the number of three types of vehicles (car, bus, and truck), total equivalent vehicles, average traffic density, average flow, average traffic speed, average acceleration in both the X and Y directions, and lane space occupancy. Specifically, the average traffic flow refers to the total number of vehicles passing through a road segment in the same direction at a given time. Lane space occupancy refers to the ratio of the road area occupied by vehicles' ground projections to the total road area in the same direction along the recorded segment. The calculation formula for lane space occupancy is given as follows:

\begin{equation}
    R_s(t) = \frac{1}{nL} \cdot \sum_{i=1}^{m} l_i 
    \label{eq:1}
\end{equation}

\noindent where \textit{L} represents the actual length of the recorded road segment, \textit{n} represents the number of lanes in the same direction, \textit{m} is the total number of vehicles in the same direction at the given time, and \textit{l} is the length of each vehicle.

\begin{figure*}[htbp]
    \centering
    \includegraphics[width=1\linewidth]{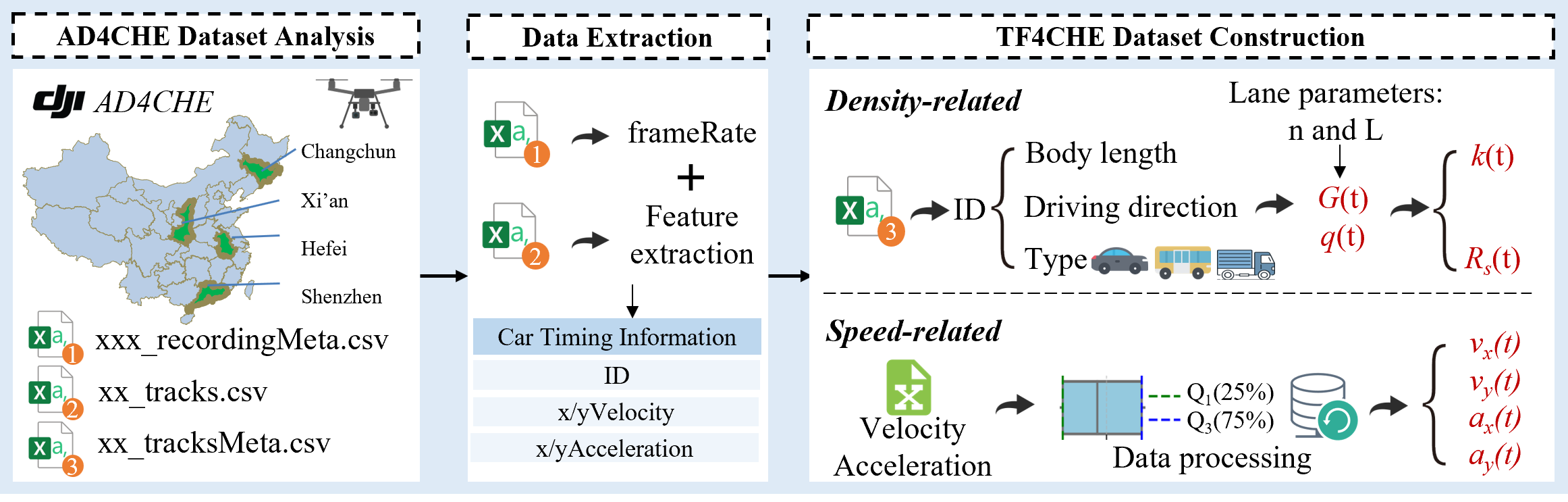}
    \caption{TF4CHE dataset construction process.Density-related parameters are primarily derived by combining vehicle-specific attributes (such as vehicle type and length) with relevant mathematical formulations, while speed-related parameters are computed through a series of data processing procedures applied to the raw data.}
    \label{fig: TF4CHE}
\end{figure*}

The average traffic speed and average traffic acceleration described in this paper represent the average instantaneous speed and acceleration of all vehicles at a given moment on the selected road segment, respectively. Traffic density refers to the total number of vehicles across \textit{n} lanes within the length of the recorded road segment and is used to describe the level of road congestion. Based on the basic definition of traffic density in the field of rail transit, its calculation formula is as follows:

\begin{equation}
    k(t) = \frac{G(t)}{nL} 
    \label{eq:2}
\end{equation}

\noindent where \textit{G}(\textit{t}) represents the total number of equivalent vehicles at a given time, reflecting the number of vehicles passing through the traffic segment at that moment. It directly reflects the magnitude of the traffic flow. The expression is provided as follows:

\begin{equation}
    G(t) = G_{\text{car}}(t) + \alpha_{\text{bus}} \cdot G_{\text{bus}}(t) + \alpha_{\text{truck}} \cdot G_{\text{truck}}(t)
    \label{eq:3}
\end{equation}

\noindent where \(G_\text{bus}(t)\) and \(G_\text{truck}(t)\) represent the number of buses and trucks at the given time, respectively, and \(\alpha\) is the conversion factor. According to the relevant regulations \cite{CJJ37-2012}, the coefficient for cars is set to 1 as the reference, while the conversion factors for buses and trucks, based on the road space occupied by each vehicle, are 2.0 and 2.5, respectively.

The specific acquisition process of the TF4CHE dataset constructed in this study is illustrated in \hyperref[fig: pipline]{Fig.3}. The process begins by analyzing the original dataset to extract detailed information from the three files for each data segment. Based on the frame rate of the recorded segment in the \texttt{xx\_recordingMeta. csv} file, time-series state information for each vehicle is extracted. Subsequently, the data in \texttt{xx\_tracks.csv} is filtered to include only the vehicle ID, the corresponding average traffic speed in the X and Y directions, and the average acceleration columns. In the \texttt{xx\_tracksMeta.csv} file, vehicle type, direction of travel, and vehicle length corresponding to each vehicle ID are extracted, and the data is divided into two major categories based on the direction of travel. Using each vehicle's type and combining Equations \eqref{eq:1}, \eqref{eq:2}, and \eqref{eq:3}, the total number of equivalent vehicles \(G(t)\), average flow \(q(t)\), traffic density \textit{k(t)}, and lane space occupancy \(R_{s}(t)\) at each time point are calculated. Finally, outliers in the speed-related parameters are removed using the interquartile range method, and the average values are calculated to obtain the average traffic speeds \(v_{x}(t)\), \(v_{y}(t)\), as well as \(a_{x}(t)\), \(a_{y}(t)\).

\subsection{Data Description}
\begin{figure}[htbp]
    \centering
    \includegraphics[width=1\linewidth]{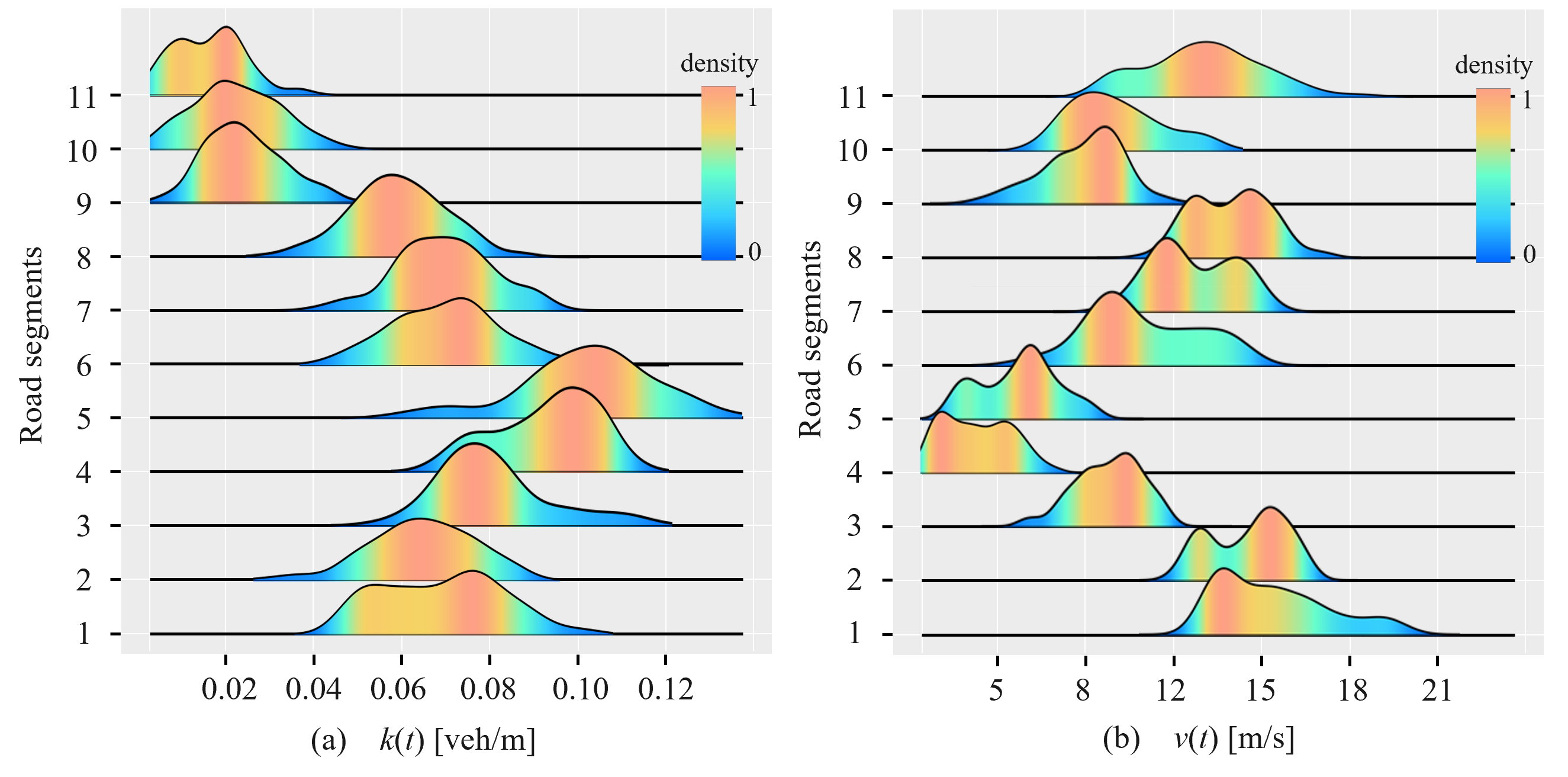}
    \caption{The KDE distribution for traffic density and average traffic speed across all road segments is shown. (a) KDE distribution for traffic density; (b) KDE distribution for average traffic speed.}
    \label{fig: KDE}
\end{figure}

This study focuses on predicting traffic density and average traffic speed in the TF4CHE dataset and uses these two parameters for congestion identification. To explore the structural characteristics and potential distribution trends of traffic density and average traffic speed data, the Kernel Density Estimation (KDE) method is used. Compared to traditional methods such as histograms, KDE provides a smoother, non-parametric estimation of the traffic density distribution, thereby more accurately revealing the data characteristics. The dataset used in this study includes 11 different road segments across 5 cities, and the KDE distribution curves for traffic density and average traffic speed in these road segments are shown in \hyperref[fig: KDE]{Fig.4}. As shown in the figure, the \(k(t)\) and \(v(t)\) data distributions in most groups exhibit a unimodal structure, generally conforming to Gaussian distribution characteristics. The overall density data are relatively high, and the average speed data are relatively low, which aligns with the characteristics of the original congested scenario dataset. Taking the road segment represented by road segment 5 as an example, the data density distribution ranges from 0.09 veh/m to 0.11 veh/m, and the average speed distribution ranges from 5 m/s to 8 m/s, which is typical of congested conditions.

\section{PPTNet Architecture\label{PPT}}
In this section, we first present the formulation of the traffic forecasting problem and outline the overall framework of the proposed model. We then elaborate on the design details, including the embedding layer, the periodic pattern extraction module, and the Transformer decoder, fully demonstrating how PPTNet effectively models the periodic features and nonlinear variations of traffic flow.
\subsection{Problem Statement}
Traffic flow prediction aim to forecast future traffic states based on observed historical traffic data. Given a historical traffic flow sequence: $\mathbf{X} = \{X_{t-T+1}, X_{t-T+2}, \dots, X_t\} \in \mathbb{R}^{T\times C}$, where \(T\) denotes the length of the historical sequence, and \(C\) represents the feature dimensionality. The prediction objective is to generate future traffic states with sequence length \(H\):
$\hat{\mathbf{Y}} = \{\hat{\mathbf{X}}_{t+1}, \hat{\mathbf{X}}_{t+2}, \dots, \hat{\mathbf{X}}_{t+H}\} \in \mathbb{R}^{H\times C}$. Typically, the historical input length \(T\) is set as multiples of the forecasting horizon \(H\), commonly ranging from \(2H\) to \(5H\) \cite{oreshkin2019n}.
\subsection{Framework Overview}
\begin{figure*}[!t]
    \centering
    \includegraphics[width=1\linewidth]{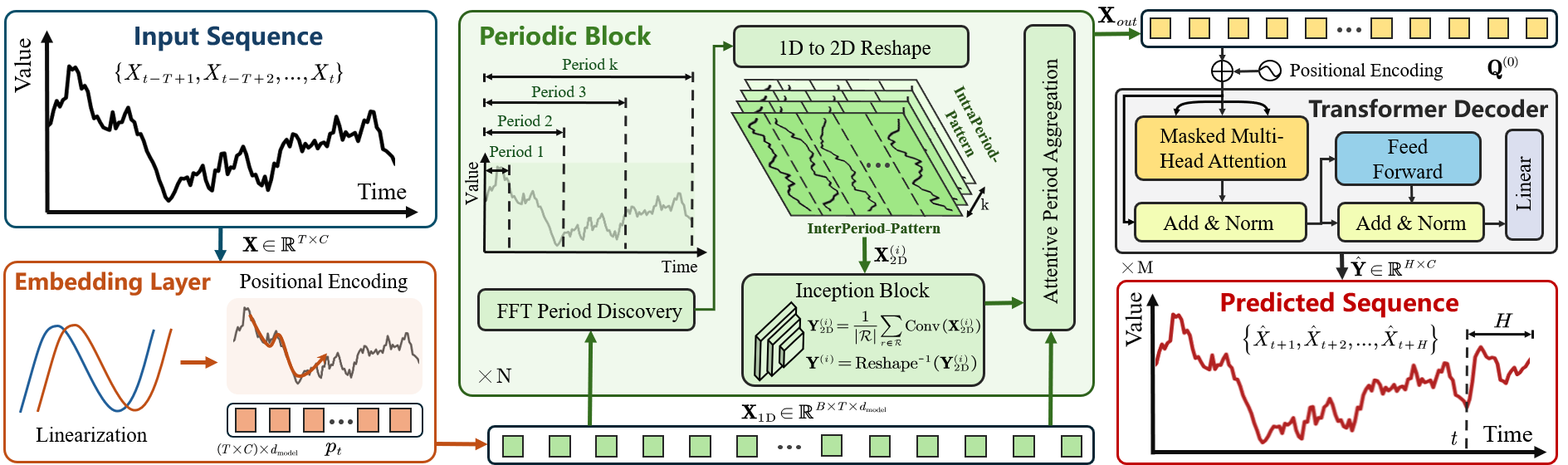}
    \caption{Architecture of PPTNet. The overall architecture consists of a Embedding layer, an encoder composed of Periodic Block that extract multi-period features via FFT and Inception convolution, and a Transformer decoder that generates multi-step predictions in the original feature space.}
    \label{fig: framework}
\end{figure*}

\hyperref[fig: framework]{Figure 5} illustrates the architecture of PPTNet. The model leverages the periodic pattern extraction capability of Periodic Blocks and the temporal modeling advantages of the Transformer architecture to address short-term traffic flow prediction tasks. The overall architecture consists of three parts: first, an embedding layer that performs a linear transformation on the original input; then, a multi-layer stacked Periodic Block, which serves as the encoder responsible for adaptively discovering primary periodic patterns from historical sequences and extracting temporal feature tables. Finally, the Transformer decoder uses the high-order feature representations extracted by the encoder to model sequence evolution through an autoregressive multi-head attention mechanism, generating traffic parameters for future time steps. 
\subsection{Embedding Layer}
Considering that the original input data $\mathbf{X} \in \mathbb{R}^{T\times C}$ has original scale differences, in order to ensure the stability of the distribution of the model input data, the input is first projected using linear transformation, and then position information is injected into each time step through position encoding. Specifically, let the embedding layer weight be \(\mathbf{W}_e \in \mathbb{R}^{C \times d_{\text{model}}}\), where \(d_{\text{model}}\) is the model feature dimension, then the calculation formula for the embedding layer is as follows:
\begin{equation}
    X_{1\mathrm{D}} = x_t W_e + p_t,\quad t = 1,2,\dots,T  
\end{equation}
where ${p}_t \in \mathbb{R}^{d_{\mathrm{model}}}$ represents the positional encoding at time step \(t\), position encoding is usually calculated using $\sin$ and  $\cos$ functions $\{p_t(2i)=\sin(t\cdot\omega_{2i})|\;p_t(2i+1)=\cos(t\cdot\omega_{2i})\}$. After passing through the embedding layer, the embedded representation is obtained as: $X_{1\mathrm{D}} \in \mathbb{R}^{T\times d_{\mathrm{model}}}$.
\subsection{Periodic Block}
\subsubsection{Multi-period Recognition based on FFT}
To address the issues of obvious periodicity, local fluctuations, and multiple frequencies in traffic flow data, Periodic Block first uses Fast Fourier Transform (FFT) to perform spectral analysis on the input sequence. For the embedded representation $\mathbf{X}_{1\mathrm{D}} \in \mathbb{R}^{B\times T\times d_{\mathrm{model}}}$, where \(B\) is the batch size, performing fast Fourier transform on the sequence along the time dimension yields the complex spectrum $\mathbf{X}_f = \mathcal{F}\{\mathbf{X}_\mathrm{1D}\} \in \mathbb{C}^{B\times(T/2+1)\times d_{\text{model}}}$. The spectrum amplitude is calculated using the amplitude averaging method as follows:
\begin{equation}
A(f)
= \frac{1}{d_{\mathrm{model}}}
  \sum_{j=1}^{d_{\mathrm{model}}}
    \bigl|\mathbf{X}_f(f,j)\bigr|\,, 
\quad
f = 0,1,\dots,\frac{T}{2}
\end{equation}
\noindent 
where $A(f)$ represents the average amplitude at frequency index $f$. To eliminate the influence of the direct current component, we set $A(0)=0$. Then select the $k$ frequencies $\{f_1, f_2, \dots, f_k\}$ corresponding to the $k$ largest amplitudes in the spectrum and estimate their periods as $p_i = \left\lfloor {T}/{f_i} \right\rfloor,\enspace i = 1,2,\dots,k.$ Where $T$ is the sampling length. Moreover, the mean amplitude $A(f_i)$ for each frequency $f_i$ is defined as the initial period weight $a_i$, ensuring that the network can adaptively capture the multiple periodic patterns inherent in the traffic data.
\subsubsection{Transformation from 1D to 2D}
After obtaining each period $p_i$, the Periodic Block performs a two-dimensional reconstruction of the input sequence. To ensure that subsequent processing uniformly covers the entire sequence, when the prediction horizon is $H$, the total required length is $T + H$. If $T + H$ is not divisible by $p_i$, the input sequence is extended with zero padding to length $L$ (with $L \ge T + H$ and $L \bmod p_i = 0$). Subsequently, the padded sequence is reshaped according to period $p_i$:
\begin{equation}
    \mathbf{X}_{2\mathrm{D}}^{(i)}
= \mathrm{Reshape}\bigl(\mathrm{Padding}(\mathbf{X}_{1\mathrm{D}})\bigr)
\;\in\;
\mathbb{R}^{B \times L \times d_{\mathrm{model}}}
\end{equation}

Each column in the 2D tensor represents the temporal variation within a single cycle, while each row reflects the variation at the same time across different cycle segments. This transformation effectively converts the originally difficult to capture 1D temporal patterns into a 2D spatial domain, allowing the subsequent 2D convolution to simultaneously extract both local and cross-cycle information.
\subsubsection{Inception Feature Extraction}
After obtaining the 2D sequence representation, the Periodic Block employs a parameter‐efficient Inception module to extract multi‐scale features from the two‐dimensional tensor $\mathbf{X}_{2\mathrm{D}}^{(i)}$. This module applies multiple 2D convolutional kernels of progressively increasing size to capture both short‐term local dynamics and larger‐scale periodic patterns. For the $i$-th periodic branch, the feature resulting from convolution with kernel size $r$ is denoted as 
$\mathbf{F}_r^{(i)} = \mathrm{Conv}_{r\times r}\bigl(\mathbf{X}_{2\mathrm{D}}^{(i)}\bigr)$, 
where $r\in\mathcal{R}$ and $\mathcal{R}$ is the set of Inception kernel sizes. The different scale convolution results are aggregated by averaging to obtain the fused 2D tensor representation as follows:
\begin{equation}
    \mathbf{Y}_{2\mathrm{D}}^{(i)}
= \frac{1}{\lvert \mathcal{R}\rvert}
  \sum_{r\in\mathcal{R}}
    \mathbf{F}_r^{(i)}.
\end{equation}

Finally, the 2D features are mapped back to the 1D temporal form by inverse remodeling. The corresponding 1D output is obtained as:
\begin{equation}
    \mathbf{Y}^{(i)}
= \mathrm{Reshape}^{-1}\bigl(\mathbf{Y}_{2\mathrm{D}}^{(i)}\bigr)
\;\in\;
\mathbb{R}^{B \times (T + H) \times d_{\mathrm{model}}}
\end{equation}

\subsubsection{Adaptive Periodic Aggregation}
Because the importance of features extracted by different periodic branches varies across contexts, the Periodic Block incorporates an adaptive period fusion module to weight and combine these branch features. The $k$ periodic branching outputs $\{\mathbf{Y}^{(1)}, \mathbf{Y}^{(2)}, \dots, \mathbf{Y}^{(k)}\}$ obtained after the above steps, are stacked along the final dimension to form the tensor $\mathbf{P}\in \mathbb{R}^{B\times (T+H)\times d_{\mathrm{model}}\times k}$. Simultaneously, each branch’s FFT amplitude weight vector is denoted by $\mathbf{a}=[a_{1}, a_{2}, \dots, a_{k}]\in \mathbb{R}^{B\times k}$. To capture the significance of each periodic branch at the current time step, the module first computes attention scores based on the mean features of each branch at the last time step:
\begin{equation}
    u_i = \frac{1}{d_{\mathrm{model}}}
      \sum_{d=1}^{d_{\mathrm{model}}}
        P\bigl(b,\,T+H,\,d,\,i\bigr),
\quad
i = 1,2,\dots,k
\end{equation}
where $b$ represents the batch index, $d$ represents the index of the model feature dimension.

The attention score vector $\mathbf{u}\in\mathbb{R}^{B\times k}$ is processed by a two-layer fully connected network to yield the attention weight vector $\boldsymbol{\alpha}=\mathrm{Softmax}(W_2\,\sigma(W_1\mathbf{u}+b_1)+b_2)\in\mathbb{R}^{B\times k}$, where $W_1$ and $W_2$ are the weights of the first and second layers, $\sigma(\cdot)$ denotes the ReLU activation function, and $b_1$ and $b_2$ are the corresponding biases. The attention weights are then multiplied element-wise with the FFT amplitude weights and normalized to obtain the final fusion weights:
\begin{equation}
    w_i \;=\; \frac{a_i \cdot \alpha_i}{\sum_{j=1}^k a_j \cdot \alpha_j}\,, 
\quad i = 1,\dots,k
\end{equation}

Using these weights to perform weighted summation of the outputs of each cycle branch, we obtain:
\begin{equation}
    \mathbf{Y}
= \sum_{i=1}^{k} w_{i}\,\mathbf{Y}^{(i)}
\;\in\;
\mathbb{R}^{B \times (T+H) \times d_{\mathrm{model}}}
\end{equation}

Finally, using residual connections, the output of the Periodic Block is obtained as $\mathbf{X}_{\mathrm{out}} = \mathbf{Y} + \mathbf{X}_{1\mathrm{D}}$. Thus, by introducing a dynamic attention mechanism, the model's ability to express periodic features in different scenarios is further improved.

\subsection{Transformer Decoder}
After processing by the multi-layer stacked Periodic Block encoder, a Transformer is employed as the decoder to capture dependencies among the $H$ future time steps. Within each decoder layer, multi-head self-attention is first applied to the input. Denote the prediction query tensor by $\mathbf{Q}_0 \in \mathbb{R}^{1\times H \times d_{\mathrm{model}}}$, which is expanded to batch size via $\mathbf{Q}^{(0)} = \mathrm{Expand}(\mathbf{Q}_0)\in\mathbb{R}^{B\times H\times d_{\mathrm{model}}}$. For the current decoder layer input $\mathbf{Q}^{(l-1)}\in\mathbb{R}^{B\times H\times d_{\mathrm{model}}}$, the learnable projection matrices yield $\mathbf{Q} = \mathbf{Q}^{(l-1)}W^Q$, $\mathbf{K} = \mathbf{Q}^{(l-1)}W^K$, and $\mathbf{V} = \mathbf{Q}^{(l-1)}W^V$, where $W^Q, W^K, W^V\in\mathbb{R}^{d_{\mathrm{model}}\times d_k}$ are linear projections, $d_k$ is the subspace dimension per head satisfying $d_{\mathrm{model}}=h\times d_k$ with $h$ attention heads. Then the attention output is as follows:
\begin{equation}
    \mathrm{Attention}(Q, K, V)
= \mathrm{softmax}\!\Bigl(\frac{QK^{\!\top}}{\sqrt{d_k}} + M\Bigr)\,V
\end{equation}
where $M\in\mathbb{R}^{H\times H}$ represents the causal mask matrix that prevents the decoder from attending to future positions during prediction: for positions $i$ and $j$, if $j>i$ then $M(i,j)=-\infty$, and otherwise $M(i,j)=0$. The outputs of the $h$ attention heads are concatenated and linearly projected back to the original model dimension, and this result is added to the input via a residual connection followed by layer normalization to produce the output of the current layer:
\begin{equation}
    \mathbf{Q}^{(l)}
= \mathrm{LayerNorm}\bigl(\mathbf{Q}^{(l-1)} + \mathrm{MHSA}(\mathbf{Q}^{(l-1)})\bigr)
\end{equation}

After $L$ decoder layers, the predicted hidden state $\mathbf{Q}^{(L)}\in\mathbb{R}^{B\times H\times d_{\mathrm{model}}}$ is obtained. It is then mapped to the final output space via a linear projection, and the output for each prediction step is given by:
\begin{equation}
    \hat{\mathbf{X}}_{t+i}
= \mathbf{q}_{t+i}^{(L)} W_{o} + b_{o}\,, 
\quad i = 1,\dots,H
\end{equation}
where $\hat{\mathbf{X}}_{t+i}$ represents the predicted traffic parameters at time $t+i$ seconds; $W_{o}\in\mathbb{R}^{d_{\mathrm{model}}\times C}$ are the projection weights; and $b_{o}\in\mathbb{R}^{C}$ are the biases. The final predicted sequence $\hat{\mathbf{Y}}\in\mathbb{R}^{H\times C}$ is then de-normalized and supplied to the subsequent fuzzy inference module.

\section{Fuzzy Inference\label{FI}}
The mapping relationship between traffic parameters and traffic states generally exhibits nonlinear behavior. Based on the traffic flow density and average traffic speed data predicted from the TF4CHE dataset, this study employs the Mamdani \cite{mamdani1975experiment} 
fuzzy inference method to calculate the probability of traffic congestion, thereby simulating the congestion level under different density and speed conditions.

\subsection{Gaussian Membership}
As shown in \hyperref[fig: KDE]{Fig. 4}, the distributions of traffic flow density and average traffic speed data are concentrated, smooth, and approximately symmetric. Therefore, a Gaussian membership function is used to describe the degree of membership of these traffic parameters. The original input data, namely traffic flow density or average traffic speed, are time-series variables in this study and are represented as follows:
\begin{equation}
    X = \{x_1, x_2, \dots, x_t\}, \quad x_i \in \mathbb{N} \label{eq:example_set}
\end{equation}

Based on the statistical characteristics of the data and domain expertise in traffic studies, traffic flow density and average traffic speed are divided into three fuzzy subsets: low, medium, and high. The Gaussian membership function corresponding to the \textit{i}-th fuzzy subset is expressed as:
\begin{equation}
    \mu_i(x) = e^{-\frac{(x - \mu_i)^2}{2\sigma^2}}, \quad i = 1, 2, \ldots, M
    \label{eq:mu_i}
\end{equation}

\noindent where $\mu_i$ represents the center of the \textit{i}-th fuzzy set, and $\sigma$ is the uniform standard deviation. $M$ denotes the number of fuzzy subsets, and in this study, $M$ = 3, corresponding to the low, medium, and high fuzzy subsets. The center of the \textit{i}-th fuzzy set is given by:
\begin{equation}
    \mu_{\text{i}} = x_{\min} + \frac{i - 1}{M - 1} \cdot (x_{\max} - x_{\min})
    \label{eq:u_i}
\end{equation}

\noindent where $x_{\min}$=$min(X)$ and $x_{\max}$=$max(X)$ represent the minimum and maximum values of the original data, respectively. To ensure that adjacent fuzzy sets have an appropriate degree of overlap, the standard deviation is determined using the following fixed-width strategy:
\begin{equation}
    \sigma = \frac{x_{\max} - x_{\min}}{2M}
    \label{eq:standard_deviation}
\end{equation}

\begin{figure}[ht]
    \centering
    \includegraphics[width=1\linewidth]{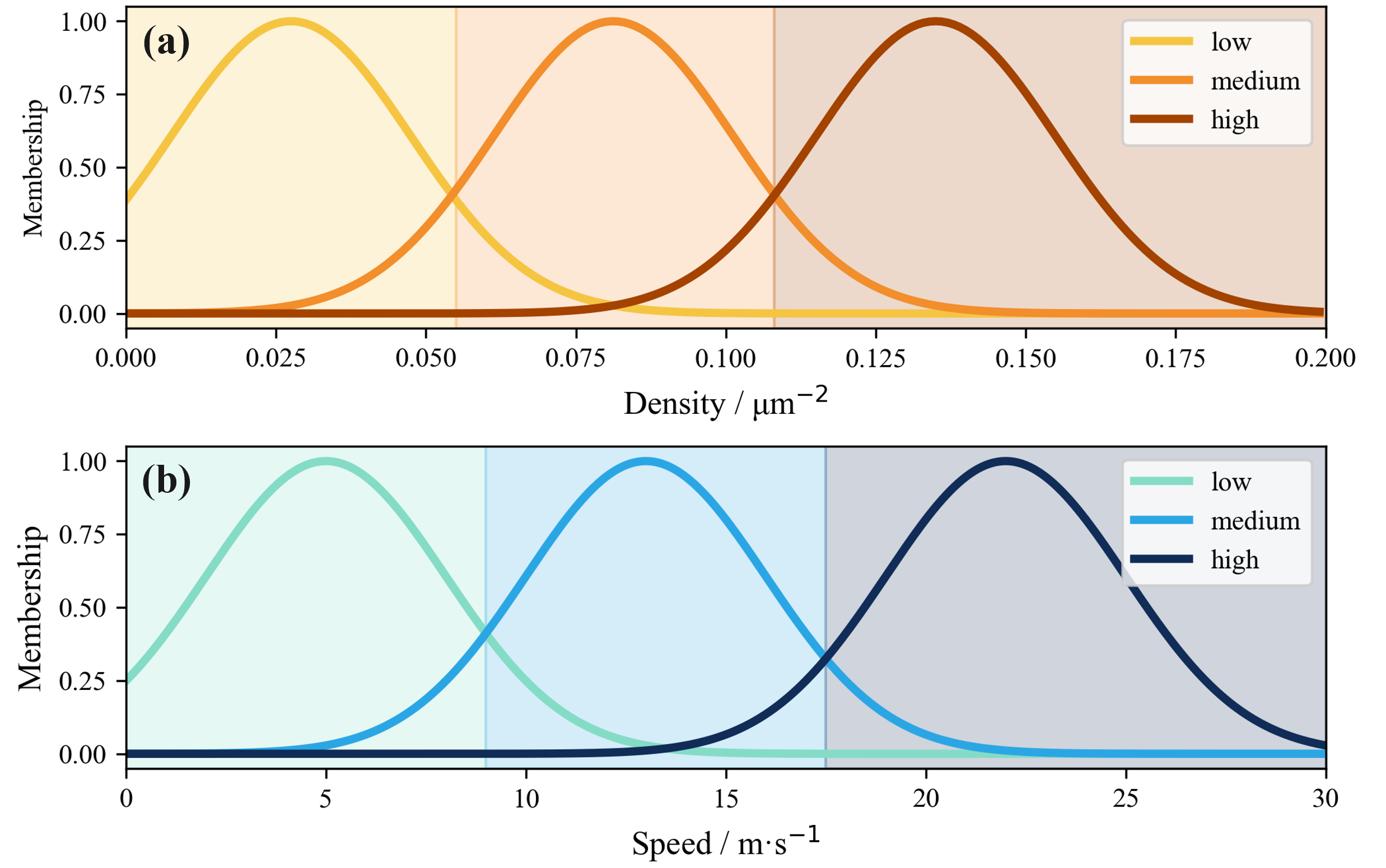}
    \caption{Gaussian membership functions for traffic flow parameters: (a) Gaussian membership function for traffic flow density; (b) Gaussian membership function for average traffic speed.}
    \label{fig: DS_Membership}
\end{figure}

The constructed Gaussian membership functions for traffic flow density and average traffic speed are shown in \hyperref[fig: DS_Membership]{Fig. 6}.

\subsection{Fuzzy Rules}
\begin{figure}[ht]
    \centering
    \includegraphics[width=1\linewidth]{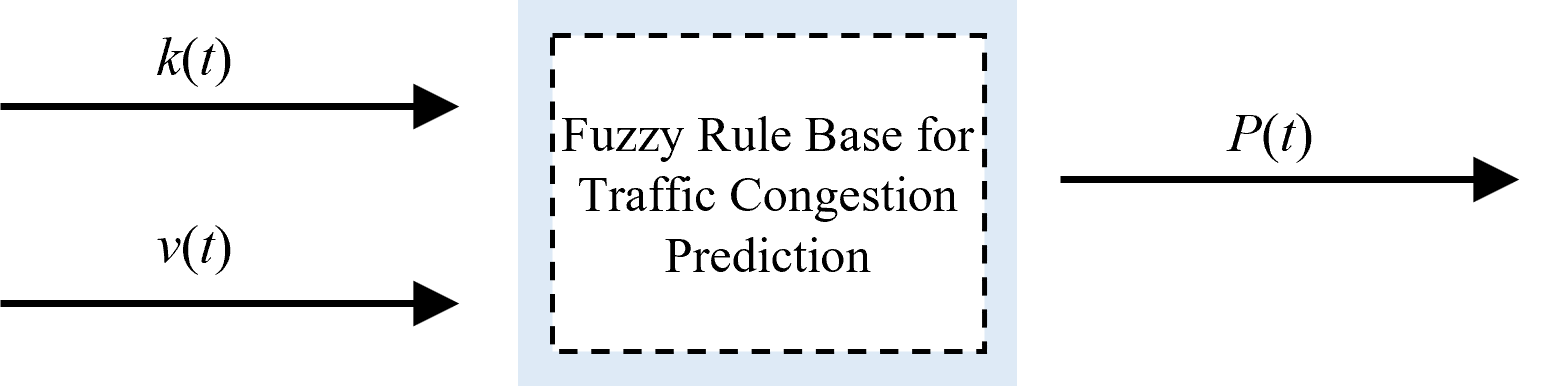}
    \caption{Fuzzy logic block diagram for traffic congestion prediction.}
    \label{fig: Fuzzy logic}
\end{figure}

Once the membership degrees of traffic flow density or average traffic speed are determined, traffic state prediction must be performed based on the fuzzy rule base. The fuzzy rule base for traffic congestion prediction constructed in this study is a dual-input, single-output form, as shown in \hyperref[fig: Fuzzy logic]{Fig. 7}.

In this study, traffic congestion probability \textit{P}(\textit{t}) levels are classified into four categories: \textit{low}, \textit{medium}, \textit{high}, and \textit{full} congestion. Both traffic flow density and average traffic speed are divided into three levels: \textit{low}, \textit{medium}, and \textit{high}. Based on the combinations of these levels, a fuzzy rule base for traffic congestion prediction is developed, as shown in \cref{tab:1}.

\begin{table}[htbp]
\caption{Fuzzy Rules for Traffic Congestion Prediction}
\label{tab:1}
\centering
\begin{tabularx}{\linewidth}{>{\centering\arraybackslash}p{0.8cm} >{\centering\arraybackslash}X}
\toprule
\textbf{Types} & \textbf{Rules} \\
\midrule
\textit{Rule 1} & \textit{If} (\textit{k}(\textit{t}) is \textit{low}) and (\textit{v}(\textit{t}) is \textit{low}) \textit{then} (\textit{P}(\textit{t}) is \textit{medium}) \\
\textit{Rule 2} & \textit{If} (\textit{k}(\textit{t}) is \textit{low}) and (\textit{v}(\textit{t}) is \textit{medium}) \textit{then} (\textit{P}(\textit{t}) is \textit{low}) \\
\textit{Rule 3} & \textit{If} (\textit{k}(\textit{t}) is \textit{low}) and (\textit{v}(\textit{t}) is \textit{high}) \textit{then} (\textit{P}(\textit{t}) is \textit{low}) \\
\textit{Rule 4} & \textit{If} (\textit{k}(\textit{t}) is \textit{medium}) and (\textit{v}(\textit{t}) is \textit{low}) \textit{then} (\textit{P}(\textit{t}) is \textit{high}) \\
\textit{Rule 5} & \textit{If} (\textit{k}(\textit{t}) is \textit{medium}) and (\textit{v}(\textit{t}) is \textit{medium}) \textit{then} (\textit{P}(\textit{t}) is \textit{medium}) \\
\textit{Rule 6} & \textit{If} (\textit{k}(\textit{t}) is \textit{medium}) and (\textit{v}(\textit{t}) is \textit{high}) \textit{then} (\textit{P}(\textit{t}) is \textit{low}) \\
\textit{Rule 7} & \textit{If} (\textit{k}(\textit{t}) is \textit{high}) and (\textit{v}(\textit{t}) is \textit{low}) \textit{then} (\textit{P}(\textit{t}) is \textit{full}) \\
\textit{Rule 8} & \textit{If} (\textit{k}(\textit{t}) is \textit{high}) and (\textit{v}(\textit{t}) is \textit{medium}) \textit{then} (\textit{P}(\textit{t}) is \textit{high}) \\
\textit{Rule 9} & \textit{If} (\textit{k}(\textit{t}) is \textit{high}) and (\textit{v}(\textit{t}) is \textit{high}) \textit{then} (\textit{P}(\textit{t}) is \textit{medium}) \\
\bottomrule
\end{tabularx}
\end{table}

\begin{figure}[ht]
    \centering
    \includegraphics[width=0.9\linewidth]{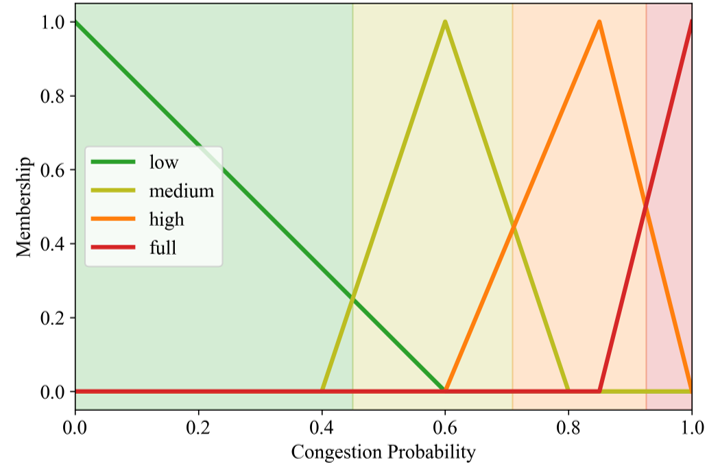}
    \caption{Triangular membership functions for traffic congestion levels.}
    \label{fig: P_Membership}
\end{figure}

The four levels of traffic congestion are represented using triangular membership functions, as illustrated in \hyperref[fig: P_Membership]{Fig. 8}.

\subsection{Defuzzification}
In this study, the centroid method is adopted to perform defuzzification of traffic congestion probability. The fuzzy output set is regarded as a mass distribution, and the defuzzified value corresponds to the coordinate of its center of gravity. First, the membership degrees $\mu_i(k)$ and $\mu_i(v)$ for each input traffic parameter are calculated. Then, the activation strength of the corresponding rules in the fuzzy rule base is computed. The activation degree of the \textit{r}-th rule is calculated using the following equation:
\begin{equation}
    \alpha_r = \min(\mu_i(k), \mu_i(v))
    \label{eq:placeholder}
\end{equation}

Each activated rule is subsequently processed through a clipping operation. The clipped output of the \textit{r}-th rule is expressed as follows:
\begin{equation}
    \mu_r^\prime(x) = \min\left(\mu(x), \alpha_r\right)
    \label{eq:mu_min}
\end{equation}

The clipped outputs of all activated rules are subsequently aggregated, as expressed by the following equation:
\begin{equation}
\mu_m(x) = \max_{r=1}^{n} \mu_r^\prime(x)
\label{eq:max_mu}
\end{equation}

Finally, the traffic congestion probability is obtained through defuzzification. The output fuzzy variable $P(t) \in [0, 1]$, and its defuzzified value is computed using the following equation:
\begin{equation}
P(t) = \frac{\int_0^1 x \cdot \mu_m(x)\, dx}{\int_0^1 \mu_m(x)\, dx}
\label{eq:defuzzification}
\end{equation}

\noindent where $x$ represents the centers of the four fuzzy sets corresponding to the defined congestion levels.

\section{EXPERIMENTS\label{ex}}
In this section, our experiments are presented as follows. First, we introduce the performance evaluation metrics and model implementation details. Then, we evaluate the proposed PPTNet prediction model on the TF4CHE dataset and validate the model accuracy and robustness through multiple ablation study. Finally, we combine the traffic flow parameters predicted by the PPTNet model with the Mamdani fuzzy inference mechanism to obtain traffic congestion probabilities, effectively validating the feasibility of the proposed method in congestion identification.
\subsection{Evaluation Metrics}
For evaluation metrics, this paper adopts three widely recognized metrics in the field of traffic prediction \cite{shu2021short}: mean absolute error (MAE), mean square error (MSE), and root mean square error (RMSE) to evaluate the prediction results. These metrics can be calculated as follows:
\begin{equation}
    \mathrm{MAE} = \frac{1}{n} \sum_{i=1}^n \bigl|y_i - \hat{y}_i\bigr|
\end{equation}
\begin{equation}
    \mathrm{MSE} = \frac{1}{n} \sum_{i=1}^n \bigl(y_i - \hat{y}_i\bigr)^2
\end{equation}
\begin{equation}
    \mathrm{RMSE} = \sqrt{\frac{1}{n} \sum_{i=1}^n \bigl(y_i - \hat{y}_i\bigr)^2}
\end{equation}
where $y_i$ represents the true value, $\hat{y}_i$ represents the predicted value, and $n$ is the sample size.

\subsection{Implementation Details\label{ID}}
The TF4CEH dataset is divided into training, validation, and test sets in a $7{:}2{:}1$ ratio, containing 17313, 5223, and 2552 sample sequences, respectively. The feature dimensions comprise the temporal feature ‘second’, vehicle count features ‘car’, ‘bus’, and ‘truck’, traffic flow features $G(t)$, $K(t)$, and $q(t)$, kinematic features $v_x(t)$, $v_y(t)$, $a_x(t)$, and $a_y(t)$, and $R_s(t)$, totaling twelve dimensions. To ensure training stability, all input features are standardized and outliers—particularly extreme values in speed-related parameters—are detected and handled using the interquartile range (IQR) method. 

\begin{table*}[!t]
\centering
\caption{Performance comparison between PPTNet and mainstream prediction models at different prediction horizons}
\vspace{-1em} 
\scriptsize
\textit{\textbf{Bold} denotes the best result, \underline{underline} denotes the best result of previous methods, $\uparrow$ denotes performance improvement, and $\downarrow$ denotes performance degradation.}
\vspace{0.3em}
\label{tab:horizon_results}
  \begin{tabularx}{0.93\textwidth}{
    >{\centering\arraybackslash}c  
    *{9}{>{\centering\arraybackslash}X}  
  }
    \toprule
    \multirow{2}{*}{Model} & \multicolumn{3}{c}{Horizon 15} & \multicolumn{3}{c}{Horizon 30} & \multicolumn{3}{c}{Horizon 45} \\
    \cmidrule(lr){2-4}\cmidrule(lr){5-7}\cmidrule(lr){8-10}
    & MAE & MSE & RMSE & MAE & MSE & RMSE & MAE & MSE & RMSE \\
    \midrule
    LSTM \cite{ma2015long}        & 0.1719 & 0.0477 & 0.2184 & 0.2517 & 0.0954 & 0.3088 & 0.2911 & 0.1179 & 0.3433 \\
    RNN \cite{belhadi2020recurrent}         & 0.1528 & 0.0316 & 0.1776 & 0.2747 & 0.0997 & 0.3157 & 0.2993 & 0.1201 & 0.3465 \\
    ConvLSTM \cite{lin2020self}    & 0.1362 & 0.0236 & 0.1535 & 0.2372 & 0.0772 & 0.2779 & 0.2747 & 0.1010 & 0.3179 \\
    Bi‐LSTM \cite{ma2021short}     & 0.1703 & 0.0410 & 0.2024 & 0.1908 & 0.0544 & 0.2331 & 0.2425 & 0.0929 & 0.3084 \\
    GRU \cite{fu2016using}         & 0.1106 & 0.0184 & 0.1356 & 0.1816 & 0.0442 & 0.2102 & 0.2060 & 0.0606 & 0.2462 \\
    CNN \cite{lecun2015deep}        & 0.1117 & 0.0166 & 0.1287 & 0.1623 & 0.0345 & 0.1857 & 0.1666 & 0.0358 & 0.1891 \\
    TCN \cite{bai2018empirical}         & 0.1036 & 0.0174 & 0.1319 & 0.0934 & 0.0151 & 0.1230 & 0.1547 & 0.0374 & 0.1934 \\
    ConvGRU \cite{shi2017deep}     & 0.0883 & 0.0119 & 0.1091 & 0.1224 & 0.0202 & 0.1420 & 0.1061 & 0.0167 & 0.1291 \\
    LSSL \cite{gu2021combining}      & 0.0859 & 0.0095 & 0.0973 & 0.0763 & 0.0078 & 0.0882 & 0.1333 & 0.0237 & 0.1539 \\
    Transformer \cite{vaswani2017attention} & 0.0946 & 0.0125 & 0.1117 & 0.0677 & 0.0069 & 0.0828 & 0.1379 & 0.0296 & 0.1721 \\
    Reformer \cite{kitaev2020reformer}    & 0.0955 & 0.0110 & 0.1049 & 0.0558 & 0.0044 & 0.0664 & 0.0711 & 0.0073 & 0.0854 \\
    FEDformer \cite{zhou2022fedformer}  & 0.0973 & 0.0087 & 0.0934 & 0.0566 & 0.0048 & 0.0693 & 0.0670 & 0.0065 & 0.0809 \\
    LSTNet \cite{lai2018modeling}     & 0.0856 & 0.0094 & 0.0972 & \underline{0.0523} & \underline{0.0040} & \underline{0.0633} & 0.0645 & 0.0059 & 0.0768 \\
    TimesNet \cite{wu2022timesnet}    & \textbf{\underline{0.0778}} & \underline{0.0085} & \underline{0.0925} & 0.0573 & 0.0048 & 0.0690 & \textbf{\underline{0.0615}} & \underline{0.0056} & \underline{0.0746} \\
    \midrule
    \rowcolor{gray!30}
    PPTNet (Ours) & 0.0821 & \textbf{0.0074} & \textbf{0.0861} & \textbf{0.0512} & \textbf{0.0033} & \textbf{0.0574} & 0.0660 & \textbf{0.0050} & \textbf{0.0709} \\
    \midrule
    \textit{Improvement} (\%) & $\downarrow$5.53 & $\uparrow$12.94 & $\uparrow$6.92 & $\uparrow$2.10 & $\uparrow$17.50 & $\uparrow$9.32 & $\downarrow$7.32 & $\uparrow$10.71 & $\uparrow$4.96 \\
    \bottomrule
  \end{tabularx}
\end{table*}

The proposed PPTNet model comprises three Periodic Blocks and a two-layer Transformer decoder; its hyperparameters are set as follows: feature dimension $64$, feedforward network dimension $128$, number of attention heads $4$, and Top-$k$ periodic patterns $6$. Training employs the AdamW optimizer with an initial learning rate of $5\times10^{-3}$, weight decay $1\times10^{-3}$, batch size $32$, dropout rate $0.2$, and a cosine annealing schedule for learning rate adjustment. All training is conducted on a workstation equipped with two NVIDIA RTX A6000 GPUs.

\subsection{Comparison of Performance}
To verify the advanced of the proposed prediction method, we compares the performance of PPTNet with 14 mainstream time series prediction models based on RNN \cite{belhadi2020recurrent}, CNN \cite{zeng2021modeling}, and Transformers \cite{vaswani2017attention}. The prediction horizon are set to 15 s, 30 s, and 45 s. \cref{tab:horizon_results} shows the comparison results between the proposed model and the baseline models. The results show that PPTNet demonstrates a significant advantage across all prediction time horizons, particularly at the 30-second prediction time horizon, where the MAE, MSE, and RMSE are 0.0512, 0.0033, and 0.0574, respectively, representing improvements of 2.1\%, 17.5\%, and 9.3\% compared to the second-best model, LSTNet, highlighting its excellent short-term prediction performance. Traditional RNN-based models (LSTM, RNN) performed poorly, mainly due to their limited ability to capture long-term periodic dependencies. Models focused on periodic modeling (such as TimesNet, FEDformer, and PPTNet) showed significant advantages in all prediction time lengths, proving the importance of periodic information for traffic flow prediction. 

\begin{figure*}[htbp]
  \centering
  \subfloat{%
    \includegraphics[width=0.48\textwidth]{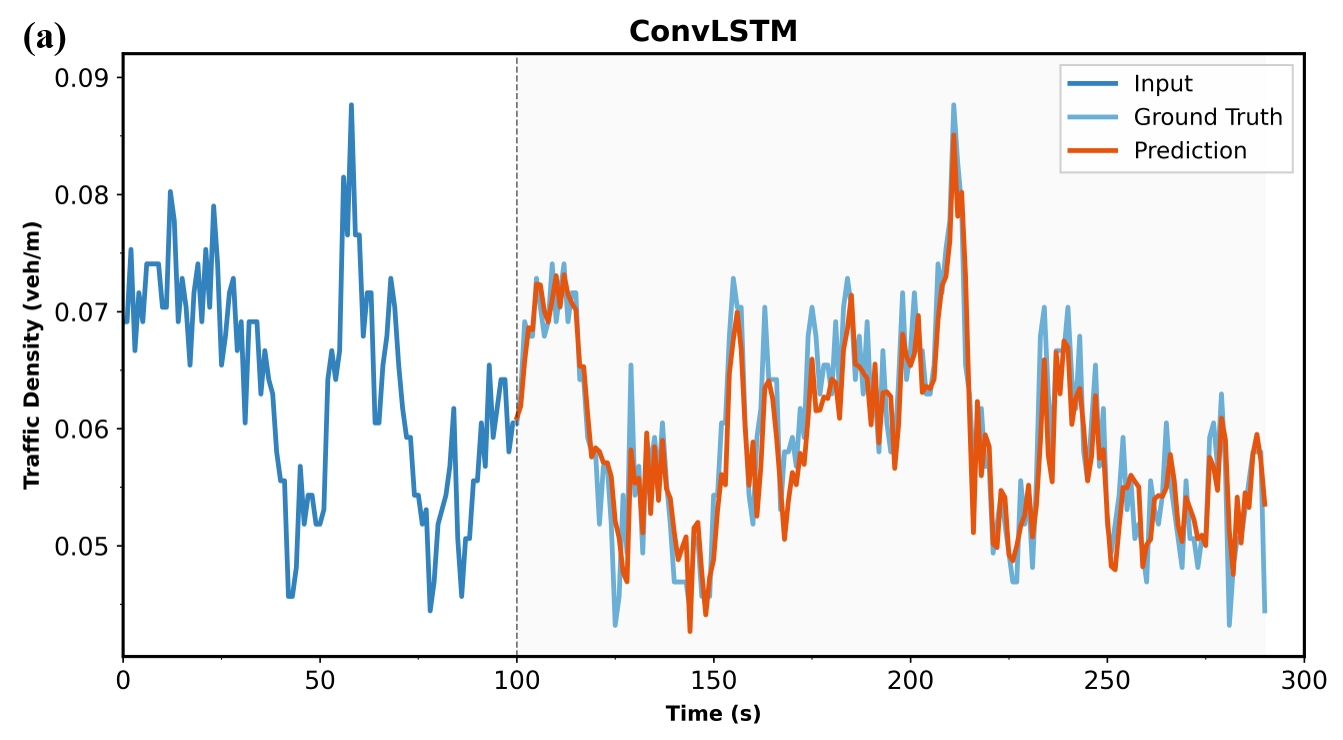}%
    \label{fig:sub1}%
  }\hfill
  \subfloat{%
    \includegraphics[width=0.48\textwidth]{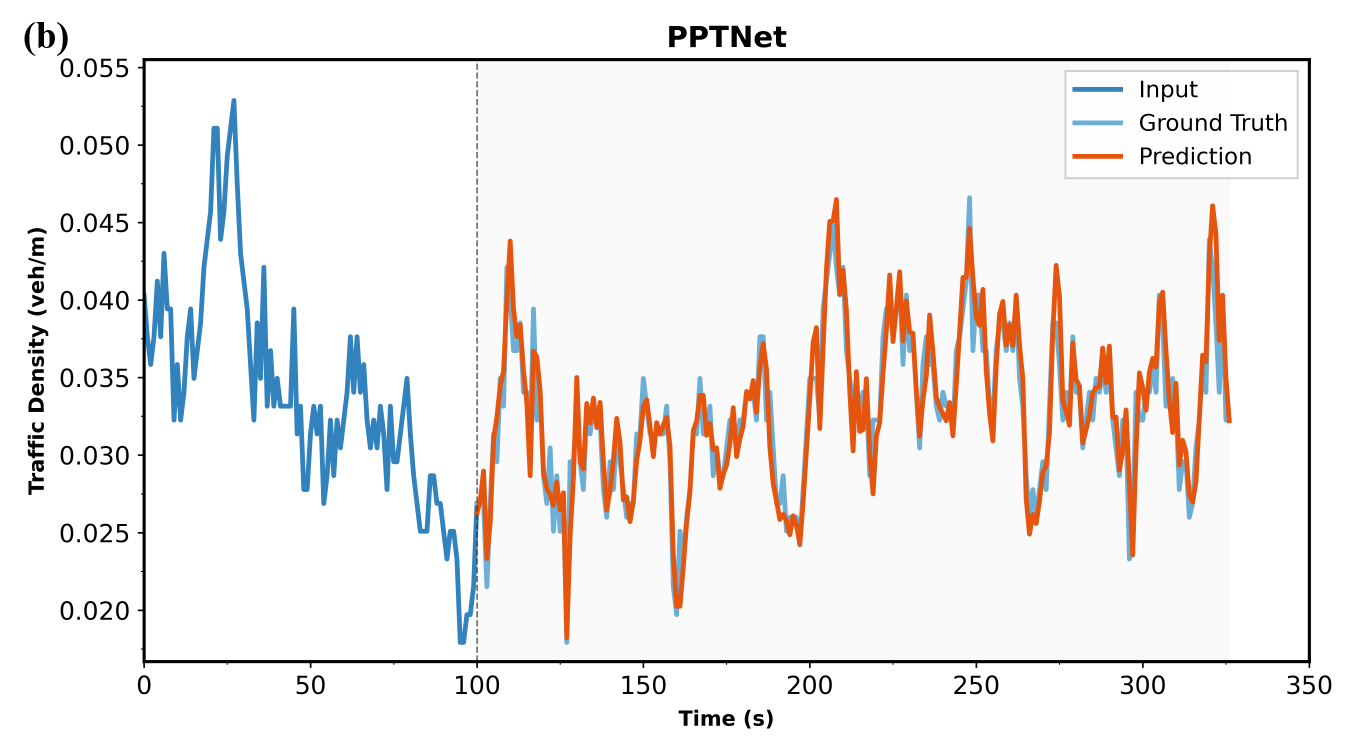}%
    \label{fig:sub2}%
  }\\[0pt]
  \vspace{-1em}
  \subfloat{%
    \includegraphics[width=0.48\textwidth]{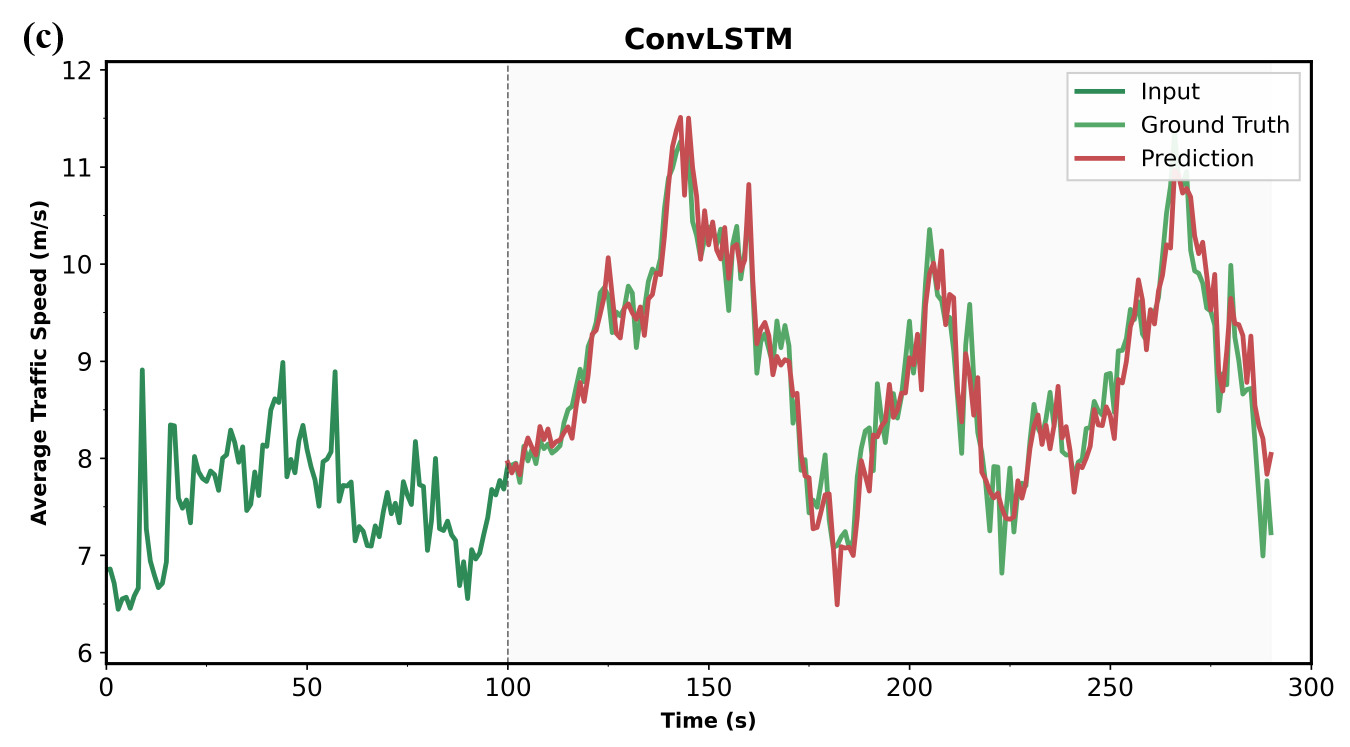}%
    \label{fig:sub3}%
  }\hfill
  \subfloat{%
    \includegraphics[width=0.48\textwidth]{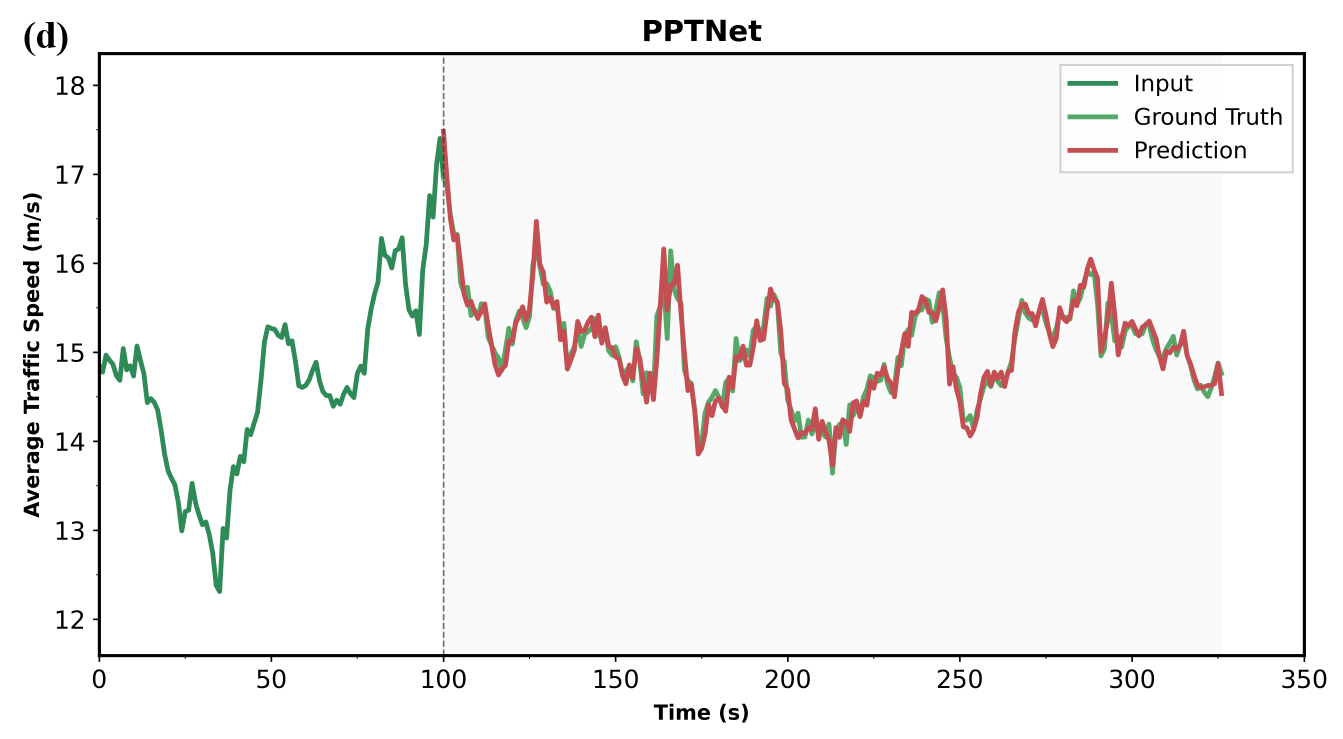}%
    \label{fig:sub4}%
  }
  \vspace{-0.5em}
  \caption{Visualization results of predictions from two different models. As shown in Figures (a) and (b), PPTNet significantly outperforms ConvLSTM in predicting density change trends, especially in accurately predicting the inflection points and peak positions of density fluctuations. As shown in Figures (c) and (d), PPTNet achieves significantly higher prediction accuracy than ConvLSTM near sudden speed changes, particularly in accurately capturing sudden speed drops and speed recovery processes.}
  \label{fig: compare}
\end{figure*}

Additionally, \hyperref[fig: compare]{Fig.9} shows a visual comparison between PPTNet and ConvLSTM \cite{lin2020self} in predicting two key parameters: traffic density and average vehicle flow speed.

\subsection{Ablation Study}
\subsubsection{Module Performance Comparison}
To evaluate the effectiveness of the periodic module and the Transformer decoder within our model, we designed three experimental configurations: one employing only the periodic module, one employing only the Transformer decoder, and one integrating both modules, as shown in \cref{tab:ablation_study}. The results demonstrate that each component, when applied in isolation, contributes to performance improvement, while the full PPTNet model yields the best overall performance. These findings validate our design rationale: the Periodic Block effectively captures the recurring patterns inherent in traffic flow data, whereas the Transformer Decoder is adept at modeling long-range dependencies within the sequence, and their combination produces a pronounced synergistic effect.

\begin{table}[!h]
  \caption{Ablation Study of Model Components}
  \label{tab:ablation_study}
  \centering
  \begin{tabular}{ccc|ccc}
    \hline
    \textbf{Baseline} & \textbf{Periodic Block} & \textbf{Transformer} & \textbf{MAE} & \textbf{MSE} & \textbf{RMSE} \\
    \hline
    $\checkmark$ & $\checkmark$ &  & 0.0710 & 0.0072 & 0.0848 \\
    $\checkmark$ &  & $\checkmark$ & 0.0600 & 0.0055 & 0.0742 \\
    $\checkmark$ & $\checkmark$ & $\checkmark$ & \textbf{0.0512} & \textbf{0.0033} & \textbf{0.0574} \\
    \hline
  \end{tabular}
\end{table}

\subsubsection{Selection of \texorpdfstring{$K$}{K} in Periodic Block}
\begin{figure}[!ht]
    \centering
    \includegraphics[width=1\linewidth]{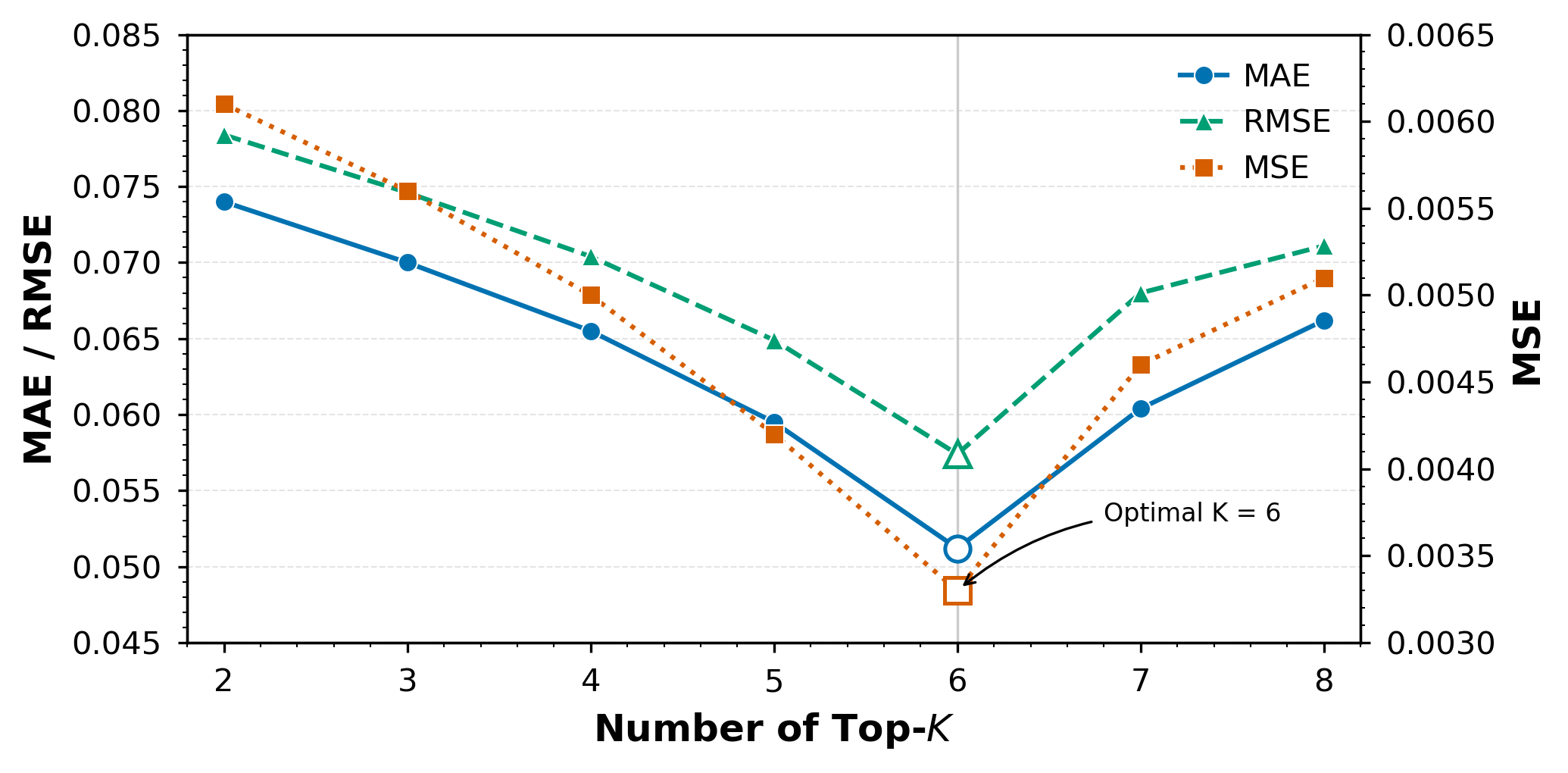}
    \caption{Impact of K in Periodic Block}
    \label{fig: topk}
\end{figure}

As illustrated in \hyperref[fig: topk]{Fig.10}, increasing the number of periods $K$ initially enhances model performance, with an optimal value of $K$ = 6 for capturing temporal features. Beyond this point, additional periods introduce noisy components that lead to overfitting. These results demonstrate that appropriately selecting the number of periodic patterns is critical for effectively modeling temporal information.

\begin{table}[!b]
  \caption{Comparison of single and multi-dimensional input}
  \label{tab:feature_comparison}
  \centering
  \begin{tabular}{l|cc}
    \hline
    \textbf{Feature Type}         & \textbf{MAE}  & \textbf{RMSE} \\ 
    \hline
    Single‐dimensional feature    & 0.1372        & 0.1463        \\ 
    Multi‐dimensional feature     & \textbf{0.0512}     & \textbf{0.0574}        \\ 
    \hline
  \end{tabular}
\end{table}

\subsubsection{Input Dimensional Comparison}
\cref{tab:feature_comparison} presents the effect of input feature dimensionality on model performance. The multi-dimensional features use the 12 feature dimensions mentioned in Section Section \ref{ID} as input. Single-dimensional features only uses density $K(t)$ as input. The experimental results reveal that incorporating multi-dimensional features substantially enhances predictive accuracy: the MAE decreases from 0.1372 with single-dimensional inputs to 0.0512 with multi-dimensional inputs, and the RMSE falls from 0.1463 to 0.0574. These findings confirm that richer feature dimensionality plays a critical role in improving forecasting performance.

\subsection{Congestion Identification Study}
\begin{figure}[ht]
    \centering
    \includegraphics[width=1\linewidth]{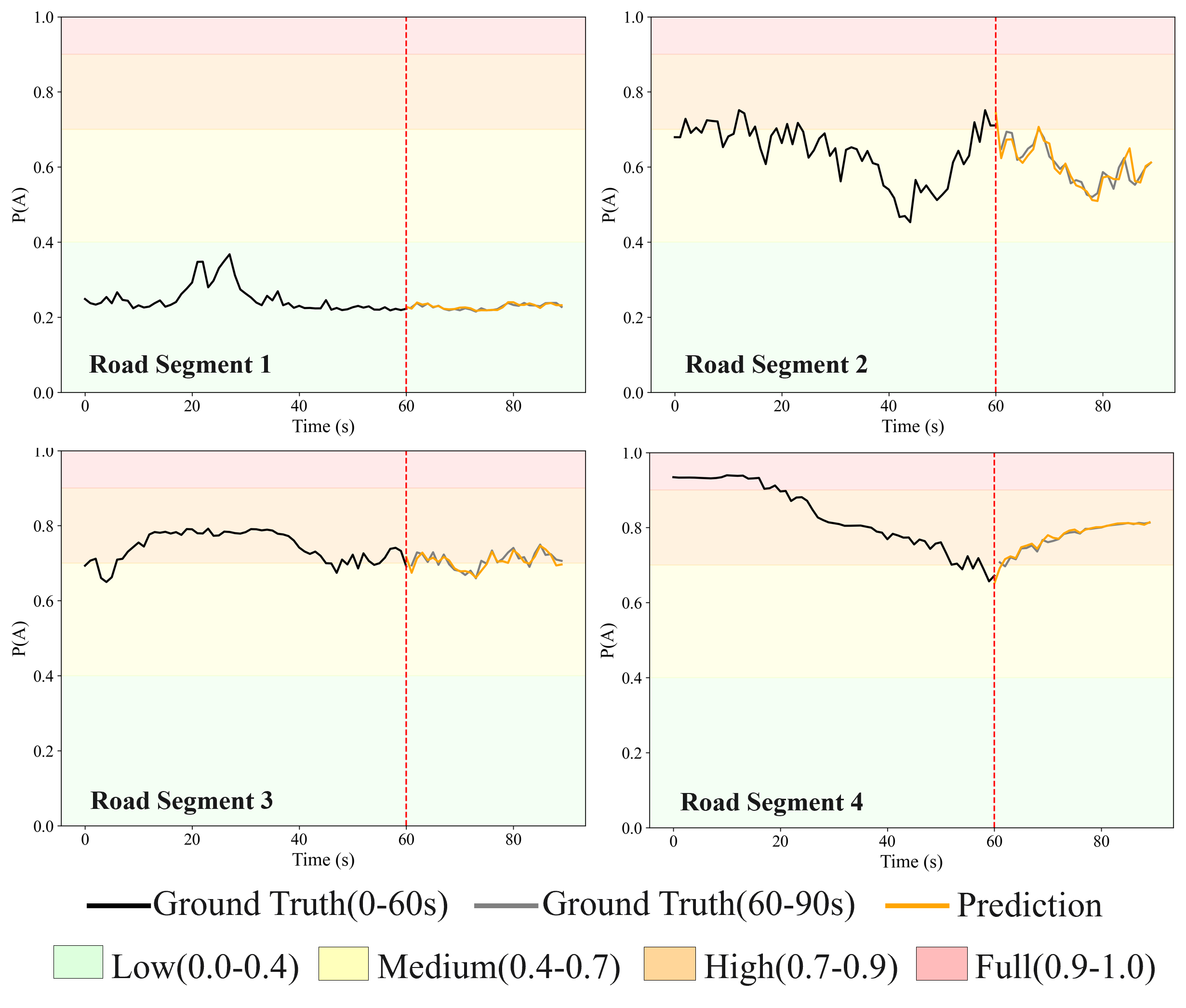}
    \caption{Predicted congestion probabilities based on PPTNet forecasts and the proposed congestion identification method.}
    \label{fig: P(A)_result}
\end{figure}

The preceding experimental analysis verifies the accuracy and robustness of the proposed traffic flow parameter prediction method. To further validate the proposed Mamdani fuzzy inference-based congestion identification approach, the 11 road segments from the TF4CHE dataset are employed for validation, with four road segments randomly selected for visualization. First, the PPTNet is used to predict traffic flow density \(k(t)\) and average traffic speed \(v_{x}(t)\) over a selected historical segment, yielding predicted values for both parameters over the next 30 s. Based on the proposed congestion identification method, the true and predicted congestion probabilities $P(t)$ over the 30 s prediction horizon are computed separately using density and speed. The comparison curves corresponding to the selected scenarios are shown in \hyperref[fig: P(A)_result]{Fig.11}.

The results of congestion probability prediction indicate that the predicted and actual congestion levels, obtained through the proposed congestion identification method, exhibit consistent categorical distributions and closely aligned numerical curves. Taking road segment 4 as an example, both the ground truth and predicted data show that the congestion probability levels are primarily distributed in the \textit{high} category after 60 seconds. These findings confirm the effectiveness and feasibility of the proposed identification approach and further demonstrate that integrating the traffic flow prediction method with the congestion identification framework can effectively forecast future traffic congestion conditions.

\section{Conclusion\label{CL}}
This paper introduces PPTNet, a hybrid traffic flow prediction model that integrates periodic patterns with a Transformer architecture, along with a fuzzy inference method for real-time congestion identification. A specialized TF4CHE dataset tailored for congested highways in China is constructed, serving as the basis for comprehensive experimental analyses. The results demonstrate that the proposed PPTNet effectively captures multi-scale periodic features and temporal dependencies inherent in traffic flow data, significantly enhancing prediction accuracy. Moreover, the developed Mamdani fuzzy inference congestion identification module reliably determines congestion states in real-time based on predicted data, exhibiting remarkable accuracy and responsiveness. Future research will explore adaptability to various traffic scenarios and further improvements in model generalization, thus advancing practical applications of intelligent transportation technologies.

{
\fontsize{8}{9}\selectfont 
\bibliographystyle{IEEEtran}
\bibliography{arxiv}
}

\end{document}